%% file: elsarticle-template.tex
\documentclass[review]{elsarticle}

\usepackage{lineno,hyperref}
\modulolinenumbers[5]

\journal{Journal of \LaTeX\ Templates}

\usepackage{longtable}

\usepackage{times}
\usepackage{soul}
\usepackage{url}
\usepackage[utf8]{inputenc}
\usepackage[small]{caption}
\usepackage{graphicx}
\usepackage{amsmath}
\usepackage{amsfonts}
\usepackage{booktabs}
\usepackage{algorithm}
\usepackage{algorithmic}

\input{def}

\begin{document}

\begin{frontmatter}

\title{THOR: Threshold-Based Ranking Loss for Ordinal Regression}


\author[mymainaddress]{Tzeviya Sylvia Fuchs \corref{mycorrespondingauthor}}
\cortext[mycorrespondingauthor]{Corresponding author}
\ead{fuchstz@cs.biu.ac.il}

\author[mymainaddress]{Joseph Keshet}

\address[mymainaddress]{Bar-Ilan University, Ramat Gan, Israel}

\begin{abstract}
In this work, we present a regression-based ordinal regression algorithm for supervised classification of instances into ordinal categories. In contrast to previous methods, in this work the decision boundaries between categories are predefined, and the algorithm learns to project the input examples onto their appropriate scores according to these predefined boundaries. This is achieved by adding a novel threshold-based pairwise loss function that aims at minimizing the regression error, which in turn minimizes the Mean Absolute Error (MAE) measure. We implemented our proposed architecture-agnostic method using the CNN-framework for feature extraction. Experimental results on five real-world benchmarks demonstrate  that the proposed algorithm achieves the best MAE results compared to state-of-the-art ordinal regression algorithms.
\end{abstract}


\end{frontmatter}


\section{Introduction} \label{sec:intro}

Ordinal regression (OR) is the prediction of an ordinal variable, where the significance of the  ordinal variable's value is due to its ordering relative to the other values. The problem setting for this task is the same as in multiclass classification, where in this case the various classes are related and their ordering is significant. The relative ordering of the labels is not captured by loss functions used for multiclass classification, such as multi-category cross entropy, which assumes independence between class labels. Therefore, one must use ordinal regression, which is usually considered as an intermediate problem between regression and classification, and may be implemented using ranking methods.

Ordinal data includes a wide range of domains, such as medical diagnosis, estimating the age of a person \cite{niu2016ordinal, chen2017using}, aesthetic \cite{liu2018constrained}, depth estimation \cite{fu2018deep}, etc., and are referred to as ordinal regression tasks. These tasks could be solved using \emph{classification-related} methods or \emph{regression-related} methods. 

Previous ordinal regression algorithms that were modified from well-known \emph{classification} algorithms are \citep{herbrich1999support}'s method of support vector learning for ordinal regression, and \cite{crammer2002pranking}'s method for ordinal regression inspired by the perceptron algorithm. More recently, Deep Neural Networks (DNNs) have been used for this task. Various works \citep{niu2016ordinal,chen2017using,cao2019rank,fu2018deep} focused on transforming the multiple-rank ordinal regression problem into binary classifiers, where every binary classifier is assigned to a specific class index, and predicts whether the ranking of the input example is higher or lower than the class index assigned to the classifier.

Alternatively, when using \emph{regression-related} methods for ordinal regression, the algorithm maps its input onto a plane, where the ranks of instances are located within intervals along that plane. In these approaches, the goal is to learn the mapping function and the boundaries (thresholds) that define the intervals. In \citep{liu2018constrained}, the authors designed a network that combined the classification-related methods and the regression-related methods: it had both a component trained for classification and a component trained for regression. The inference was determined by the component trained for classification.

In this work, we argue that classification methods are not always suited for the OR task, since they usually ignore the ordering between classes, and because their objective function is not one that typically suits the goal of OR. In regression tasks, on the other hand, it is difficult to find the appropriate boundaries between labels. We propose a novel ordinal regression algorithm, solved by a regression approach that uses a predefined set of boundaries; i.e., the boundaries are set and not learned, while the mapping function learns to assign instances with appropriate scores along the plane accordingly. This is an extension to the work of \citep{fuchs2017spoken}, where a ranking algorithm predicts the relevance score of an instance in accordance with and relative to a single threshold value that has been set in advance. Our proposed method, which we denote by THresholded Ordinal Regression (THOR) is architecture-agnostic and can be used with any deep neural network.

This paper is organized as follows. Section \ref{sec:previous_work} reviews the literature of ordinal regression. 

We  present our proposed method in Section \ref{sec:method}, and in Section \ref{sec:experiments} we show experimental results and  various applications of our proposed method. Finally, concluding remarks and future directions are discussed in Section \ref{sec:conclusions}.

\section{Related Work}
\label{sec:previous_work}

As mentioned previously, ordinal regression could be solved using classification related methods or regression related methods. One of the better known classification methods is Ordinal Regression with CNN (OR-CNN) \citep{niu2016ordinal}, in which the authors transformed an ordinal regression problem with $K$ classes into $K-1$ binary classifiers; the $k$-th binary classifier predicted whether the rank of an instance was greater than $k$. Thus, the label of the first class was converted into $[0, ..., 0]$, the second class was converted into $[1, ..., 0]$, the third class was $[1,1, ..., 0]$ and so on. 

\begin{figure*}
  \centering
  \begin{minipage}[b]{0.2\textwidth}
  \centering
    \includegraphics[width=\textwidth]{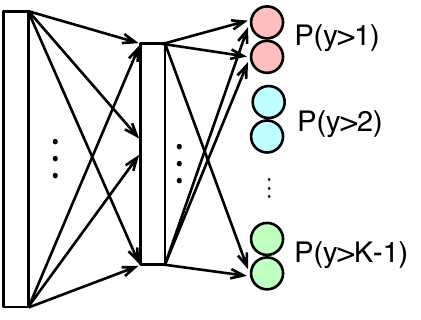}
    \text{(a) OR-CNN}
  \end{minipage}
   \hfill
   \begin{minipage}[b]{0.2\textwidth}
     \includegraphics[width=\textwidth]{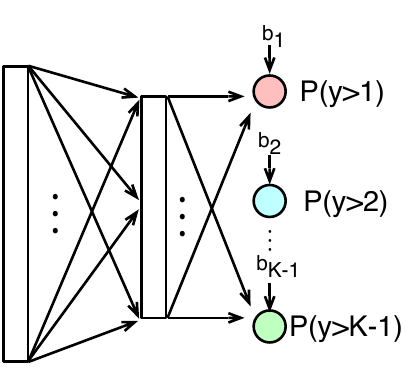}
     \text{(b) CORAL}
   \end{minipage}
   \hfill
   \begin{minipage}[b]{0.2\textwidth}
     \includegraphics[width=\textwidth]{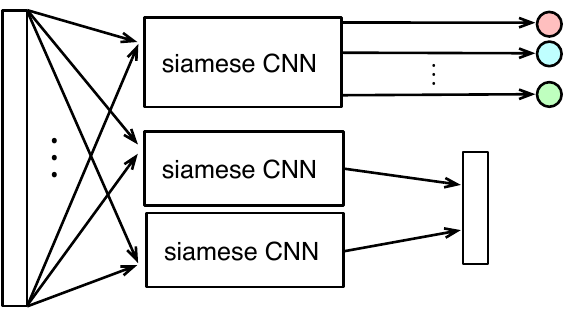}
   	\text{(c) CNNPOR}
   \end{minipage}
   \hfill
   \begin{minipage}[b]{0.2\textwidth}
     \includegraphics[width=\textwidth]{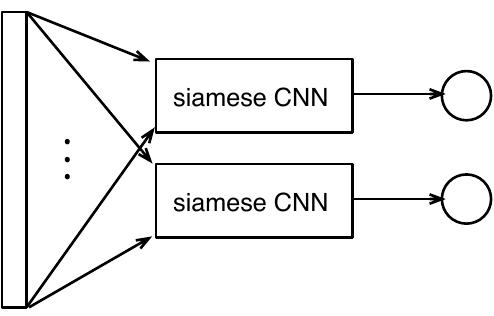}
\text{(d) THOR} 
   \end{minipage}
  \caption{Ordinal regression methods' architecture comparison. (a): OR-CNN, which has $K-1$ binary classification problems. (b): CORAL, where every one of the $K-1$ binary problems has an independent bias. (c): CNNPOR, containing a classification component and a ranking component. (d): THOR, which consists of a Siamese network with a pairwise thresholded ranking loss.}
  \label{fig:diagrams}
\end{figure*}

In OR-CNN, each example is fed-forward into a CNN, at the end of which are $K-1$ independent classifiers. The final prediction $\hat{y}$ is calculated by
\begin{equation}
    \hat{y} = \sum_{k=1}^{K-1}[f_k(x) =1] +1,
\end{equation}

where $f_k(x)$ is the $k$-th output of a DNN for an input $x$. The OR-CNN architecture is depicted in Figure \ref{fig:diagrams}(a).

However, in the above OR-CNN method, the various binary classifier branches $f_k(x)$ are not necessarily monotonically decreasing, since they are independent and could be inconsistent. That is, the network could predict the input $x$ to be both greater than $k+1$ and smaller than $k-1$, simultaneously. 

Therefore, \citep{cao2019rank} proposed the COnsistent RAnk Logits (CORAL) method. CORAL is a method similar to OR-CNN, with the difference of there being $K-1$ binary tasks that share the same weight parameters, but have independent bias units. In CORAL, each example is fed-forward into a CNN, at the end of which is a single output. The $K-1$ independent bias units create the $K-1$ binary tasks. The final prediction $\hat{y}$ is calculated similar to the method in OR-CNN. In their work, the authors prove that their method solves the consistency problems of the binary classifiers. The CORAL architecture is depicted in Figure \ref{fig:diagrams}(b).

The regression related methods are considered to be \textit{threshold approaches}. These approaches assume that a function maps the input instances to a real line, and the ranks of the instances are intervals along the line. These methods treat discrete labels as continuous numerical values, in which the adjacent classes are equally distant. They also assume that the $K$ classes are separable and ordered in a unique direction. That is, to classify input instances into $K$ ranks, the algorithm would typically use $K-1$ thresholds. The threshold approaches therefore need to learn the mapping function and the boundaries of the intervals. Methods that use the thresholds approach include SVOR \citep{chu2007support} and GPOR \citep{chu2005gaussian}. 

In \citep{liu2018constrained}, the authors propose a Convolutional Neural Network with Pairwise regularization for Ordinal Regression (CNNPOR), which combines the classification and regression methods into a single architecture. They use a triplet Siamese CNN to process their pairwise dataset. Denote the three components of the Siamese CNN as $f_1$, $f_2$ and $f_3$, where $f_1$ is used for classification, and $f_2$ and $f_3$ are used for regression. More specifically, suppose the system is given two examples $x^{(i)}$ and $x^{(i+1)}$ which belong to classes $i$ and $i+1$ respectively. The two examples are first fed into $f_1$, whose output has $K$ separate units with a logistic regression loss, similar to multiclass classification. Then,  $x^{(i)}$ and $x^{(i+1)}$ are each simultaneously fed into $f_2$ and $f_3$, whose outputs are fed into a pairwise hinge loss. The system's final loss combines the logistic regression loss and the pairwise hinge loss as follows
\begin{equation}
    \ell_{\mathrm{CNNPOR}} = l_1 + C \times l_2,
\end{equation}
where $l_1$ is the logistic regression loss, $l_2$ is the pairwise hinge loss, and $C$ is a hyper-parameter set to 1. The CNNPOR architecture is depicted in Figure \ref{fig:diagrams}(c).

\section{Method}
\label{sec:method}

Denote by $x \in \X$ an input instance to be classified, and $y \in \Y$ its label, where $\Y = \{1, 2, 3, ..., K\}$ are all possible ranks. Our goal is to find a mapping function $f:\X \to \reals$ that would yield a value from which we can infer on the input's rank. This is done by a set of boundaries $B = \{b_0, b_1, b_2, ..., b_K \}$ whose values are set in advance. An input whose value falls within the segment $[b_{i-1}, b_{i}]$ will be predicted to belong to class $i$.

We now describe  our  loss function  formally. Consider two examples drawn at random from two consecutive classes, i.e., $x^{(i)} \in \X^{(i)}$ and $x^{(i+1)} \in \X^{(i+1)}$. The probability that the mapping function $f$ assigns the two examples to their correct segments in $\reals$ is 
\begin{align}
    ACC = \mathbb{P} ~ &\{ b_{i-1} < f(x^{(i)}) < b_i
    ~~ \wedge ~~  b_{i} < f(x^{(i+1)}) < b_{i+1} \}, \label{eq:acc}
\end{align}

\MakeLowercase{Where} $\wedge$ is the logical conjunction symbol. Thus, our goal is to find the parameters of the function $f$ that would maximize the $ACC$ value for a given set of boundaries $B$. Equivalently we find the parameters of function $f$ that minimize the error defined as
\begin{align}
Err &= 1-ACC \nonumber \\
 &= \mathbb{P} ~ \{ f(x^{(i)})< b_{i-1} ~~\vee~~ f(x^{(i)}) > b_i ~~\vee~~  f(x^{(i+1)})< b_{i} ~~\vee~~  f(x^{(i+1)}) > b_{i+1} \} \nonumber\\
 &= \mathbb{E} ~ \Big[ \indicator {  f(x^{(i)})< b_{i-1} } + \indicator {  f(x^{(i)}) > b_i} + \indicator {  f(x^{(i+1)})< b_{i} } + \indicator {  (x^{(i+1)}) > b_{i+1}} 
 \Big], \label{eq:err_function}
\end{align}

where $\vee$ is the logical disjunction symbol, and $\indicator{\pi}$ is the indicator function, that equals 1 if the predicate $\pi$ holds true and 0 otherwise. 

\begin{figure}[t]
 \centering
 \includegraphics[width=0.55 \linewidth]{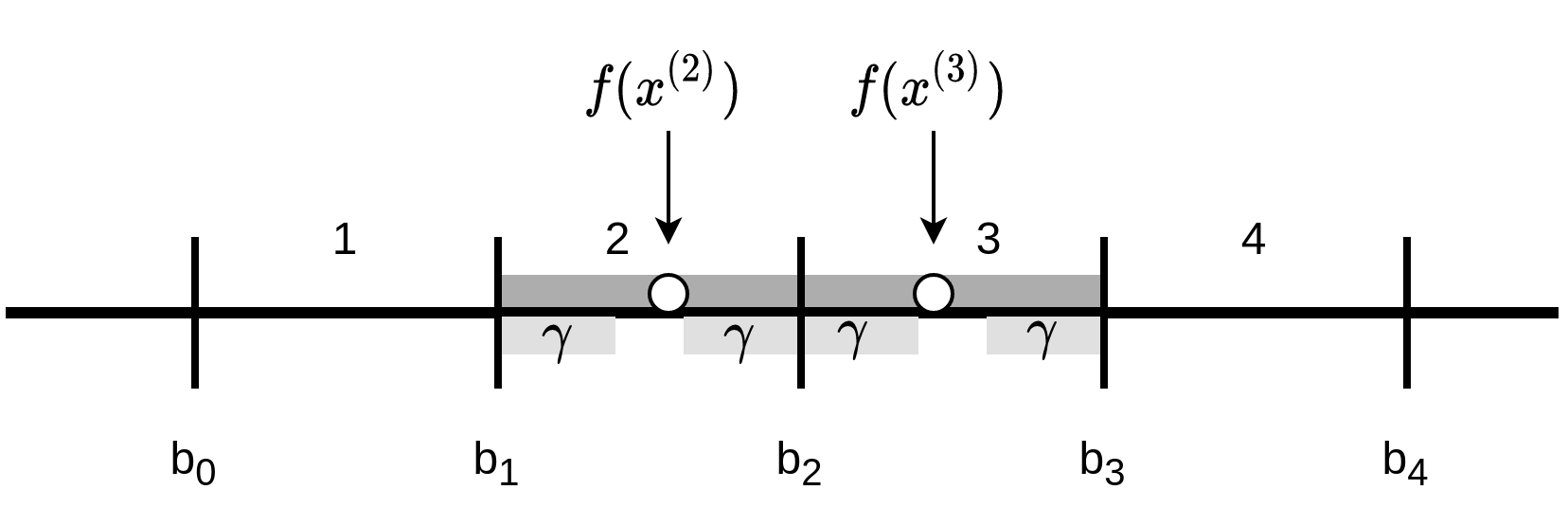}
 \caption{ Demonstrating the ACC measure in Eq. \eqref{eq:acc} and \eqref{eq:loss}. In the example, the score of $f(x^{(2)})$ must be between the boundaries $b_1 + \gamma$ and $b_2 - \gamma$, while the score of $f(x^{(3)})$ must be between $b_2 + \gamma$ and $b_3 - \gamma$.}
 \label{fig:thresholds}
\end{figure}

Since the equation in \eqref{eq:err_function} cannot be minimized directly, we replace it with a surrogate loss function, specifically a convex upper bound to the error function that is easy to minimize, as follows: 
\begin{align}
    Err &\le \mathbb{E} ~ \Big[
    [\gamma + b_{i-1} - f(x^{(i)})]_+  + [ \gamma - b_i + f(x^{(i)})]_+ \nonumber \\
&~~~~~~~~~~~~~~+~~~~~ [\gamma + b_i - f(x^{(i+1)})]_+ + [\gamma - b_{i+1} + f(x^{(i+1)})]_+ \Big] \label{eq:loss}
\end{align}

where $[\pi]_+ = \max\{\pi, 0\}$, and $\gamma > 0$ is the margin. The upper bound in Eq. \eqref{eq:loss} holds true since $\indicator{\pi < 0} \le [\gamma-\pi]_+$. Adding the margin $\gamma$ makes the problem that $f$ is trying to solve more difficult; i.e., the boundaries of example $f(x^{(i)})$ are no longer $b_{i-1}$ and $b_{i}$, but rather $b_{i-1} + \gamma$ and  $b_{i} - \gamma$, yielding a slimmer range for correct $f(x^{(i)})$ prediction, and creating a larger gap between classes. We illustrate this idea in Figure \ref{fig:thresholds}. The margin $\gamma$ is discarded during inference.

We present an algorithmic implementation aimed at minimizing the loss function derived from Eq. \eqref{eq:loss}, namely,  

\begin{multline}
    \ell(x^{(i)}, x^{(i+1)}; b_{i-1}, b_i, b_{i+1}) =  
    [\gamma + b_{i-1} - f(x^{(i)})]_+  + [ \gamma - b_i + f(x^{(i)})]_+ \\   + [\gamma + b_i - f(x^{(i+1)})]_+ + [\gamma - b_{i+1} + f(x^{(i+1)})]_+ \label{eq:final_loss}
\end{multline} 

\section{Experiments}
\label{sec:experiments}

We evaluate our proposed approach on five benchmarks: classifying images according to the decade they belong to, classifying the severity of medical data, classifying photo aesthetics, and age estimation according to images and speech. All images were resized to 256 × 256 pixels, and randomly cropped to 224 × 224 for data augmentation, similar to \citep{liu2018constrained}. The speech data was cut to segments of 0.5 sec each, and is represented by the Short-Time Fourier Transform (STFT) features, extracted using the \texttt{librosa} package \citep{librosa}. The features were computed on a 20 ms window, with a shift of 10 ms. 

Our algorithm architecture, THOR, is depicted in Figure \ref{fig:diagrams}(d). Since we are using the pairwise approach, at every iteration a random pair of instances with adjacent labels $x^{(i)}$ and $x^{(i+1)}$ is fed-forward into a Siamese network. Each component of the Siamese network has a single output, $f(x^{(i)})$ and $f(x^{(i+1)})$ respectively, which are in turn fed into the loss function in Eq. \eqref{eq:final_loss}.

More specifically, the mapping function $f$ is a CNN with an input layer of size $3 \times 224 \times 224 $. THOR is architecture-agnostic, and any CNN may be used. We demonstrate our results using VGG19\footnote{\texttt{https://github.com/pytorch/vision/blob/master/torchvision/models/vgg.py}}, ResNet18 and ResNet34\footnote{\texttt{https://github.com/pytorch/vision/blob/master/torchvision/models/ResNet.py}}. Since the given task has been converted to a regression problem and may need the ability of expressing a wide range of values, we slightly modified the VGG19, ResNet18 and ResNet34 architectures by adding two additional fully connected layers to their ends. We denote these modified architectures as VGG19*, ResNet18* and ResNet34*.

THOR requires $K+1$ thresholds, as the first and last labels (class $1$ and class $K$) require lower and upper boundaries as well, respectively. The threshold values $\{b_0, b_1, ... b_K \}$ were set to be $\{-1, 0, 1, ... K\}$, and the margin was set to be $\gamma = 0.5$. These values were tuned on the validation set. 

For the image datasets, all deep networks were fine tuned from the pretrained ImageNet model \citep{simonyan2014very}. The learning rate of all layers, except for the last fully connected layers, was set to 0.0001. Similar to \citep{liu2018constrained}, the learning rate of the last fully connected layers was set to 0.001. For the speech dataset, the network's parameters were initialized randomly. We optimized using Stocastic Gradient Descent (SGD). 

We used classification accuracy and Mean Absolute Error (MAE) as our performance measures. The accuracy is defined as:
\begin{equation}
Accuracy = \frac{1}{|S|} \sum_{x_t \in S}\indicator{\hat{y}_t = y_t},
\end{equation}
 where  $S$ is the test set, $y_t$ is the ground truth label of instance $x_t$,  and $\hat{y}_t$ is its predicted label. This is a hit-or-miss measure for testing, and not to be confused with the ACC measure in Eq. \eqref{eq:acc}, which we introduced in order to define our regression problem setup.

MAE is defined as:
\begin{equation}
    MAE = \frac{1}{|S|} \sum_{x_t \in S}{|\hat{y}_t - y_t|}.
\end{equation}

Conceptually, the difference between the two measures is that \emph{accuracy} assesses the classification quality whereas \emph{MAE} assesses the estimation quality. That is, \emph{accuracy} evaluates the algorithm's performance according to whether it classified correctly or not, regardless of how wrong the misclassified examples are. \emph{MAE}, on the other hand, does not measure whether the algorithm was correct or incorrect in its classification, but rather measures the prediction's distance from the true label. In this work we emphasize the importance of the \emph{MAE} measure, especially when dealing with large $K$ values, where the difference between adjacent classes is negligible and the \emph{accuracy} measure is almost meaningless. For example, in the age estimation task, when trying to estimate the age of a person in an image from within a range of 15 to 60, predicting that an image of a 30 year old is 32 years old has little significance. On the other hand, predicting that the same image is of a 50 year old is a significant error, but the classification loss would treat both types of mistakes evenly. We thus argue that minimizing \emph{MAE} suits OR tasks betten than minimizing accuracy. In fact, our loss function in \eqref{eq:final_loss}, which aims at minimizing the regression function's prediction error, uses a two-part margin ranking loss rather than a classification loss. Therefore, it can be assumed that once converged, the algorithm would minimize MAE as well.

\subsection{Datasets}
\label{ssec:datasets}

\begin{figure}[t]
 \centering
 \includegraphics[width=0.7\linewidth]{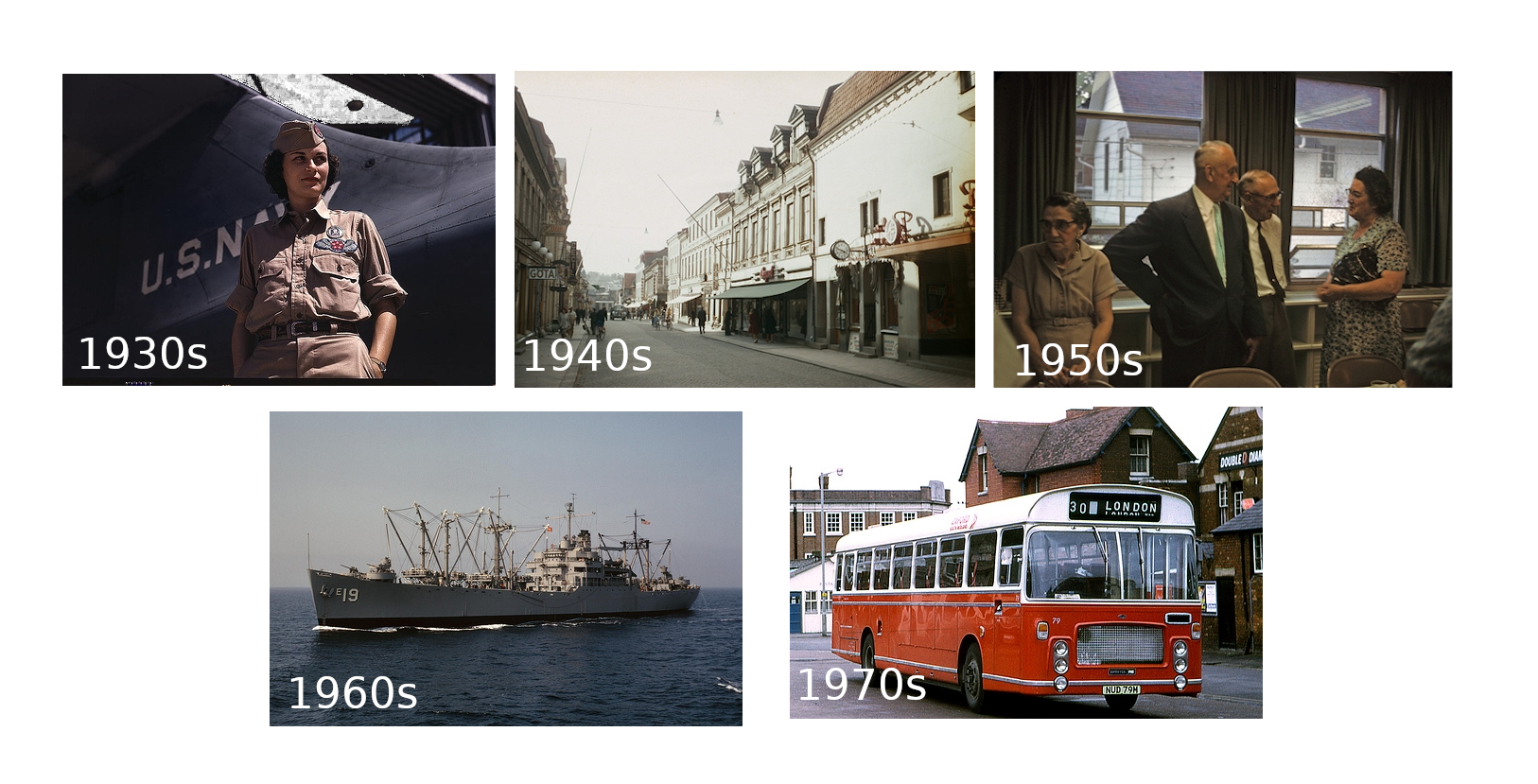}
 \caption{Examples from the Historical Color Images Dataset. }
 \label{fig:historic}
\end{figure}

\paragraph{Historical Color Images Dataset.}
\label{sssec:historical}
The historical color image dataset \citep{palermo2012dating} is a benchmark
to evaluate algorithms predicting when a historical color
image was photographed in the decade scale. The dataset contains five ordinal categories, from the 1930s to the 1970s, where each category has 265 images. Similar to  \citep{palermo2012dating}, we used 210 images for training, 5 images for validation, and 50 images for testing. Figure \ref{fig:historic} shows examples from the dataset.

\paragraph{Diabetic Retinopathy Detection.}
\label{sssec:kaggle}
The Diabetic Retinopathy (DR) Detection dataset\footnote{ \texttt{www.kaggle.com/c/diabetic-retinopathy-detection}} is a dataset released by kaggle, and consists of high resolution fundus images. They are used to predict one of five levels of diabitic retinopathy: no DR, mild, moderate, severe, and proliferative DR. The dataset consists of 17,563 pairs of images, where every pair are the left and right eye images of a single patient. The images were preprocessed as proposed in \citep{graham2015kaggle}. The images were split with a 60/20/20 ratio for training, validation and testing, respectively.  Since the number of images in each class is uneven, we used 400 randomly chosen images from each class for training and validation, and approximately 140 images from each class for testing.

\paragraph{Aesthetic Visual Analysis (AVA).}
\label{sssec:ava}
The AVA dataset \citep{avaref} is a large-scale database for aesthetic visual analysis containing more than 250,000 photos downloaded from the photography contest website dpchallenge.com. Every image is rated from 1 to 10. Similar to \citep{zhu2020deep}, for this task we averaged and rounded the scores each image received. We only used images with average scores between 3-7, as there was not enough data beyond this range. The images were split with a 60/20/20 ratio for training, validation and testing, respectively. To avoid the uneven amount of images in each class, we used 250 randomly chosen images from each class for training and validation, and 150 images from each class for testing.

\paragraph{UTKFace Dataset.}
\label{sssec:utk}

\begin{figure}[t]
 \centering
 \includegraphics[width=0.75\linewidth]{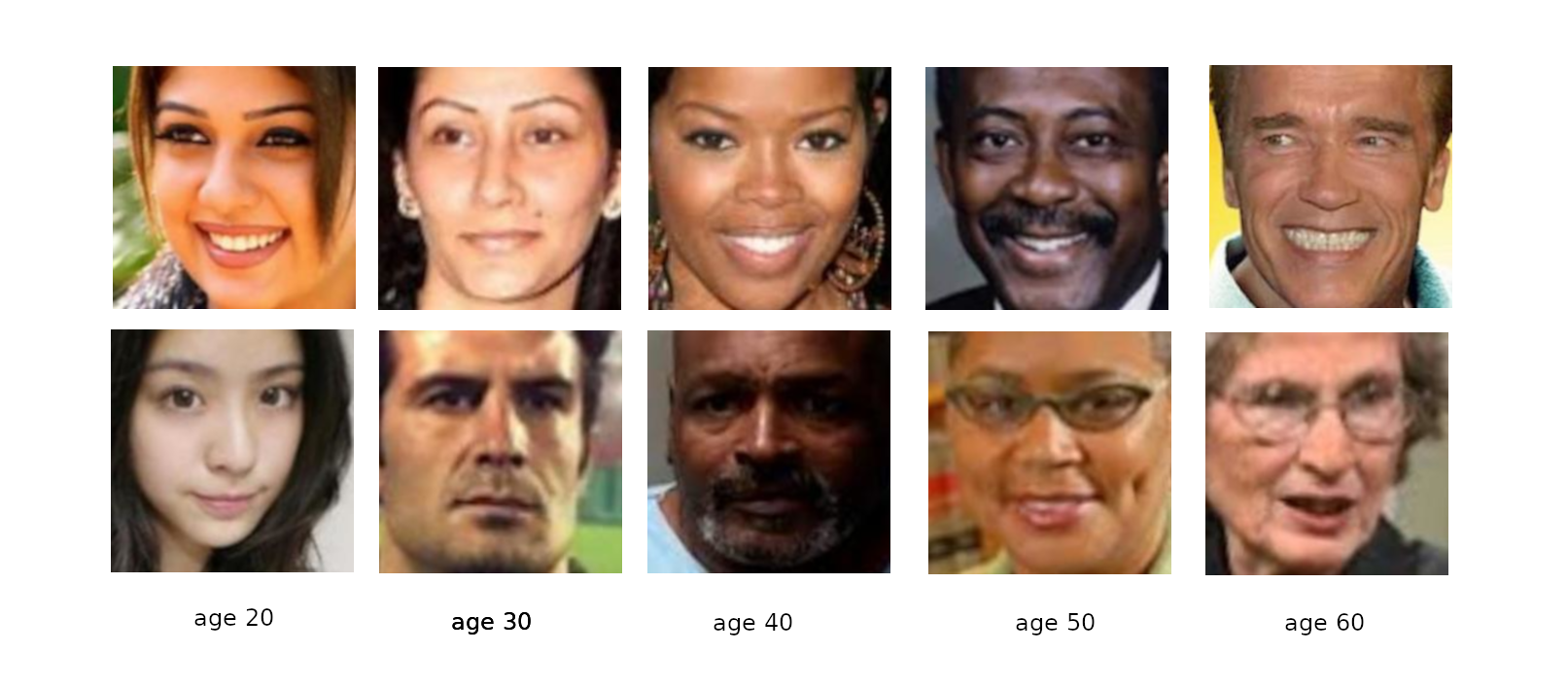}
 \caption{Examples from the UTKFace dataset. }
 \label{fig:utk}
\end{figure}

UTKFace dataset  \citep{zhifei2017cvpr} is a large scale aligned and cropped face dataset, with over 20,000 face images, and ages ranging from 0 to 116. For some experiments, we used the images of people aged 20, 30, 40, 50, and 60, with 170 images for every age category in the training and validation set, and approximately 60 images for every age category in the test set. We chose labels with a 10 year age gap between one another because otherwise the differences between the classes might be too subtle to detect. We denote this subset of the UTKFace dataset as UTKFace-5, since it includes 5 classes. For other experiments, we used all images of people from ages 0 to 68. For this setup, we refer to the dataset simply as UTKFace. Figure \ref{fig:utk} shows examples from the dataset.

\paragraph{TIDIGITS Dataset.}
\label{sssec:tidiits}
TIDIGITS \citep{leonard1993tidigits} is a dataset of read speech, with 326 speakers pronouncing 77 digit sequences. The speakers consist of both men, women, boys and girls, from ages 7 to 59. TIDIGIT's training set was split with a 90/10 ratio for training and validation, and TIDIGIT's test set was used for testing. All audio files were split to segments of 0.5 sec for convenience, and we use up to 170 speech segments for each category. As before, to ensure a sufficient gap between the age groups, only data with labels 7, 15, 25, 35, and 44 have been used. We denote this subset of the TIDIGITS dataset as TIDIGITS-5. In experiments with which the entire range of speakers has been used, we refer to the dataset simply as TIDIGITS.

\paragraph{IMDB-WIKI Dataset.}
\label{sssec:imdb}
The IMDB-WIKI dataset \citep{Rothe-IJCV-2018,Rothe-ICCVW-2015} is a large dataset of images with gender and age labels. The images are of celebrities, with their photos, name and age automatically extracted from IMDB and Wikipedia. The dataset contains 460,723 images and could be at times extremely inaccurate, with some images containing erroneous labels, and with some images not containing a face at all. For this dataset, we used the images of people between the ages 15 and 60, since outside of this range the data was relatively sparse.

\subsection{Results}
\label{ssec:results}

\subsubsection{Evaluation with a small number of tasks}

We compare our method with CNN-POR \citep{liu2018constrained}, OR-CNN \citep{niu2016ordinal} and CORAL \citep{cao2019rank} which have been presented in Section \ref{sec:previous_work}. Each one of these methods used a different specific CNN architecture, as described in their corresponding papers. To avoid introducing empirical bias, and for fair comparison, we implement all algorithms using the same CNN architectures. We present our main results in Table \ref{table:vgg19}, \ref{table:ResNet18} and Table \ref{table:ResNet34} for VGG19*, ResNet18* and ResNet34* respectively. In all tables, the best results are marked in bold, whereas the second best results are underlined.  It seems that while our method does not usually have the highest accuracy values, it is often second best. More importantly, as expected, our method always achieves best MAE values, showing that even when its predictions are not accurate - the distance from the correct label is smaller than in the other algorithms. Interestingly, the high accuracy results of CNN-POR may be due to its use of the logistic regression loss, which would naturally maximize accuracy. Our algorithm, on the other hand, is best at minimizing MAE due to its use of a ranking loss.

\begin{table*}[ht]
\centering
\scalebox{0.8}{
 \begin{tabular}{l c c c c c c c c}
 \toprule
  & \multicolumn{2}{c}{CNN-POR \ignore{\citep{liu2018constrained}}} & \multicolumn{2}{c}{OR-CNN\ignore{\cite{niu2016ordinal}}} & \multicolumn{2}{c}{CORAL\ignore{\cite{cao2019rank}} } & \multicolumn{2}{c}{THOR} \\ [0.5ex] 
\cline{2-3}  \cline{4-5} \cline{6-7} \cline{8-9}
VGG19* & Accuracy & MAE  & Accuracy & MAE & Accuracy & MAE & Accuracy & MAE \\
 \midrule
 Historical & \textbf{0.536}  &  \underline{0.748}  &  0.332 & 0.936 &  0.352 & 1.224  &  \underline{0.384} &  \textbf{0.768} \\ 
 retinopathy &  \textbf{0.468} &  0.782  &  0.447 & \underline{0.708} & 0.366  &  1.034 & \underline{0.464}  &  \textbf{0.699} \\ 
 AVA-5 & 0.\underline{325}  & 1.057   & 0.308  & \textbf{0.936} & \textbf{0.368}  & 1.121  &  0.264 & \underline{0.981} \\ 
 UTKFACE-5 &  \textbf{0.45}  &  \underline{0.811}  & 0.332  & 0.918 & 0.364  & 1.135  & \underline{0.425}  & \textbf{0.693} \\ 
 TIDIGITS-5 & \textbf{0.371} &  \underline{1.048}  & 0.234  & 1.079 & \underline{0.302}  & 1.427  & 0.255  & \textbf{0.968} \\ 
  \bottomrule
\end{tabular}}
\centering
\caption{\label{table:vgg19} Accuracy and MAE results on 5 datasets using the VGG19* architecture}
\end{table*}

\begin{table*}[ht]
\centering
\scalebox{0.8}{
 \begin{tabular}{l c c c c c c c c}
 \toprule
  & \multicolumn{2}{c}{CNN-POR \ignore{\cite{liu2018constrained}}} & \multicolumn{2}{c}{OR-CNN\ignore{\cite{niu2016ordinal}}} & \multicolumn{2}{c}{CORAL\ignore{\cite{cao2019rank}} } & \multicolumn{2}{c}{THOR} \\ [0.5ex] 
\cline{2-3}  \cline{4-5} \cline{6-7} \cline{8-9}
ResNet18* & Accuracy & MAE  & Accuracy & MAE & Accuracy & MAE & Accuracy & MAE \\
 \midrule
 Historical & \textbf{0.536} & \underline{0.848}  & 0.200 & 1.200 & 0.340 & 1.308 & \underline{0.488} & \textbf{0.720} \\ 
  retinopathy & \underline{0.430} & 1.049 & 0.200 & 1.193 & 0.370 & \underline{0.965} & \textbf{0.460} & \textbf{0.709} \\ 
  AVA-5 & \underline{0.335}& 1.128 & 0.200 & 1.200 & \textbf{0.380} & \underline{1.026} & 0.308 & \textbf{0.968} \\ 
 UTKFACE-5 & \textbf{0.436} & \underline{0.904} & 0.200&1.200 & 0.325&1.196 & \underline{0.411}&\textbf{0.789} \\ 
 TIDIGITS-5 & \textbf{0.355}&\underline{1.049} &  0.200&1.200 & 0.200&1.200 & \underline{0.268}&\textbf{0.973} \\ 
  \bottomrule
\end{tabular}}
\centering
\caption{\label{table:ResNet18} Accuracy and MAE results on 5 datasets using the ResNet18* architecture}
\end{table*}

\begin{table*}[ht]
\renewcommand{\arraystretch}{1}
\centering
\scalebox{0.8}{
 \begin{tabular}{l c c c c c c c c} 
 \toprule
  & \multicolumn{2}{c}{CNN-POR \ignore{\cite{liu2018constrained}}} & \multicolumn{2}{c}{OR-CNN\ignore{\cite{niu2016ordinal}}} & \multicolumn{2}{c}{CORAL\ignore{\cite{cao2019rank}} } & \multicolumn{2}{c}{THOR} \\ [0.5ex]
\cline{2-3}  \cline{4-5} \cline{6-7} \cline{8-9}
ResNet34* & Accuracy & MAE  & Accuracy & MAE & Accuracy & MAE & Accuracy & MAE \\
 \midrule
 Historical & \textbf{0.524}&\underline{0.872}  & 0.200&1.172 & 0.344&1.296 & \underline{0.480}&\textbf{0.748} \\ 
  retinopathy & 0.420&0.870  & \underline{0.424}&\underline{0.750}  & 0.369&1.035  &  \textbf{0.460}&\textbf{0.735} \\ 
AVA-5 & 0.324 &\underline{1.085} &  0.200 & 1.201 & \textbf{0.353}& 1.132 & \underline{0.339}& \textbf{0.965} \\
UTKFACE-5 & \textbf{0.482}&\underline{0.821} & 0.200&1.200 & 0.329&1.232 & \underline{0.457}&\textbf{0.675} \\ 
TIDIGITS-5 & \textbf{0.307}&1.142 &  0.200&\underline{1.200} & 0.200&\underline{1.200} & \underline{0.301}&\textbf{0.933} \\ 
 \bottomrule
\end{tabular}}
\centering
\caption{\label{table:ResNet34} Accuracy and MAE results on 5 datasets using the ResNet34* architecture}
\end{table*}

\begin{figure}[t]
 \centering
 \includegraphics[width=0.6\linewidth]{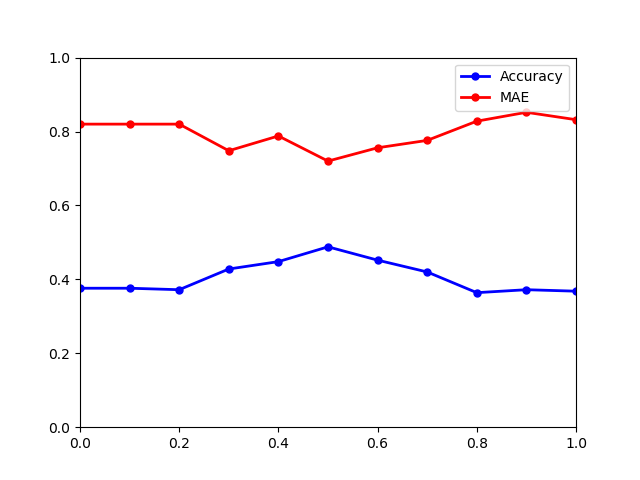}
 \caption{Accuracy and MAE results for various $\gamma$ values, evaluated using THOR on the historical color images dataset. }
 \label{fig:gamma}
\end{figure}

In all experiments, we set the margin to be $\gamma = 0.5$ after tuning it on the validation set. To further understand its effect, we plotted the accuracy and MAE results for various values of $\gamma$ within the range of 0 to 1 (with a 0.1 step size). We demonstrate its effect on the historical color images dataset, as shown in Figure \ref{fig:gamma}. It is clear that the accuracy and MAE results improve the closer $\gamma$ is to the center of the defined range, meaning that the algorithm performs best when it encourages examples to be as far as possible from its boundaries.

Since CNN-POR seems to achieve high accuracy results, and THOR is best with the MAE measure, we considered combining the two algorithms. The combined algorithm would consist of the Siamese network and thresholded ranking loss, as in THOR, and would have an additional network with tied weights for classification, similar to the classification component in CNN-POR. Notice that this new architecture could be evaluated in two ways: the inference could be performed using only the classification network, or using the output value of only the regression network. Results are in Tables \ref{table:ResNet18_cnn_thor} and \ref{table:ResNet34_cnn_thor} for ResNet18* and ResNet34* respectively. Most results did not improve over the best results found in Tables \ref{table:ResNet18} and \ref{table:ResNet34}. Interestingly, results inferred through the classification network yield higher accuracy results, whereas results inferred through the regression network  yield better MAE results.

 \begin{table}[ht]
 \centering
  \begin{tabular}{l c c c c} 
  \toprule
   & \multicolumn{2}{c}{Classification} & \multicolumn{2}{c}{Regression} \\ [0.5ex] 
 \cmidrule{2-3}  \cmidrule{4-5} 
  ResNet18* & Accuracy & MAE  & Accuracy & MAE  \\
  \midrule
  Historical & \textbf{0.544} & 0.840  &  0.492 & 0.796  \\ 
  retinopathy & 0.457 & 0.875  &  0.449 & 0.803 \\ 
 AVA-5 &  0.327 & 1.180  & 0.339  & 1.06 \\
 UTKFACE-5 & \textbf{0.454} & 0.896  & 0.414  & 0.85 \\ 
 TIDIGITS-5 & \textbf{0.364} &  1.058 & 0.348  & \textbf{0.922} \\ 
  \bottomrule
 \end{tabular}
 \centering
 \caption{\label{table:ResNet18_cnn_thor} Results of algorithm that combines CNN-POR's classification network with THOR's siamese network, using ResNet18*. Results under the ``classification'' column refer to predicting $\hat{y}$ according to the classification network. Results under the ``regression'' column refer to predicting $\hat{y}$ according to the regression network. Results that are better than those achieved in Table \ref{table:ResNet18} are in bold.}
 \end{table}

 \begin{table}[ht]
 \centering
  \begin{tabular}{l c c c c} 
  \toprule
   & \multicolumn{2}{c}{Classification} & \multicolumn{2}{c}{Regression} \\ [0.5ex] 
 \cline{2-3}  \cline{4-5} 
  ResNet34* & Accuracy & MAE  & Accuracy & MAE  \\
  \midrule
  Historical & \textbf{0.552}  & 0.816   & 0.504  & 0.777   \\ 
  retinopathy & 0.443 \ignore{delete} & 0.853 \ignore{delete}  &  0.453 \ignore{delete} & 0.879 \ignore{delete} \\ 
 AVA-5 &  0.332 & 1.690  & 0.323  & 1.056 \\
 UTKFACE-5 &  0.468 & 0.796  & 0.479  & 0.700 \\ 
 TIDIGITS-5 & \textbf{0.320} & 1.14  & 0.321  & 0.955 \\ 
  \bottomrule
 \end{tabular}
 \centering
 \caption{\label{table:ResNet34_cnn_thor} Results of algorithm that combines CNN-POR's classification network with THOR's siamese network, using ResNet34*. Results under the ``classification'' column refer to predicting $\hat{y}$ according to the classification network. Results under the ``regression'' column refer to predicting $\hat{y}$ according to the regression network. Results that are better than those achieved in Table \ref{table:ResNet34} are in bold.}
 \end{table}

\subsubsection{Evaluation with a large number of tasks}

In all the above experiments the number of classes $K$ was at most 5. To evaluate our algorithm's performance on datasets with a larger $K$ value, we applied it on the age estimation task. Thus, we measure the  performances of CNNPOR, OR-CNN and CORAL using the following datasets: 

\begin{enumerate}
    \item UTKFace, this time using the images of people between the ages 0 and 68, since outside of this range the data was relatively sparse. We use approximately 100 images for every age category.
    
    \item IMDB-WIKI, using images of people between the ages 15 and 60. Similar to the UTKFace dataset, for every category we used approximately 100 images.
    
    \item TIDIGITS, this time using the images of people between ages 7 and 59, which is the entire range provided in this dataset. 
    
\end{enumerate}

The prediction of a person's exact age is clearly a more difficult task.  Results are in Tables \ref{table:ResNet18_full} and \ref{table:ResNet34_full} for ResNet18* and ResNet34* respectively. While none of the algorithms seems to be the best on all datasets, our algorithm achieved the best MAE results for 2 out of the 3 datasets; for the IMDB-WIKI dataset, which is extremely noisy, it only achieved second-best MAE results. Regarding accuracy, CORAL seems to perform somewhat better than the rest, though clearly not always. Thus, Tables \ref{table:ResNet18_full} and \ref{table:ResNet34_full} demonstrate that our algorithms can be used for a large $K$ as well.

\begin{table}[H]
\centering
 \begin{tabular}{l c c c c } 
 \toprule
 ResNet18* & CNN-POR  & OR-CNN & CORAL  & THOR \\ [0.5ex] 
 \midrule
IMDB-WIKI & \underline{0.034}/10.782 & 0.033/\textbf{8.658} & 0.024/13.284   & \textbf{0.040}/\underline{8.979} \\ 
UTKFACE & 0.062/10.720 &  0.027/12.659& \textbf{0.093}/\underline{6.896} & \underline{0.073}/\textbf{6.047} \\ 
TIDIGITS &  \textbf{0.063}/\underline{7.103} & 0.031/8.250 & \underline{0.054}/9.317 & 0.056/\textbf{6.085} \\ 
 \bottomrule
\end{tabular}
\centering
\caption{\label{table:ResNet18_full} Accuracy and MAE results on datasets with large $K$ value, using the ResNet18* architecture. Results are presented as accuracy/MAE. }
\end{table}

\begin{table}[H]
\centering
 \begin{tabular}{l c c c c } 
 \toprule
 ResNet34* & CNN-POR & OR-CNN & CORAL  & THOR \\ [0.5ex] 
 \midrule
IMDB-WIKI & \underline{0.036}/10.344 & 0.033/\textbf{8.661} &  0.029/11.243 & \textbf{0.041}/\underline{8.975}\\ 
UTKFACE & 0.063/10.076 &  0.022/13.008 & \textbf{0.101}/\underline{6.111} &  \underline{0.083}/\textbf{5.711}\\ 
TIDIGITS & \underline{0.057}/\underline{7.390}  & 0.028/8.255 & \textbf{0.058}/8.129 & 0.069/\textbf{6.052} \\ 
 \bottomrule
\end{tabular}
\centering
\caption{\label{table:ResNet34_full} Accuracy and MAE results on datasets with large $K$ value, using the ResNet34* architecture. Results are presented as accuracy/MAE.}
\end{table}

\section{Conclusions}
\label{sec:conclusions}
In this work, we introduced a new loss function for regression-based ordinal regression, in which all boundaries are set and fixed in advance and do not need to be learned. The mapping function learns to assign instances with scores that are relative to the the predefined boundaries. Results suggest that our pairwise method and loss function yield MAE values that are in most cases better than previous works' results. 

In future work, we plan to improve on the accuracy results as well. Additionally, preliminary experiments in which the boundaries $B$ were defined as learned parameters did not perform as well as when they were set in advance, an interesting fact which requires additional research.


\bibliography{mybibfile}

\end{document}

%% file: def.tex
\newcommand{\indicator}[1]{\ensuremath{\mathbb{I}\{#1\}}}

\newcommand{\X}{\ensuremath{\mathcal{X}}}
\newcommand{\Y}{\ensuremath{\mathcal{Y}}}

 \newfont{\msym}{msbm10}
\newcommand{\reals}{\mbox{\msym R}}

\newcommand{\ignore}[1]{}

%% file: elsarticle-template.bbl
\begin{thebibliography}{10}
\expandafter\ifx\csname url\endcsname\relax
  \def\url#1{\texttt{#1}}\fi
\expandafter\ifx\csname urlprefix\endcsname\relax\def\urlprefix{URL }\fi
\expandafter\ifx\csname href\endcsname\relax
  \def\href#1#2{#2} \def\path#1{#1}\fi

\bibitem{niu2016ordinal}
Z.~Niu, M.~Zhou, L.~Wang, X.~Gao, G.~Hua, Ordinal regression with multiple
  output cnn for age estimation, in: Proceedings of the IEEE conference on
  computer vision and pattern recognition, 2016, pp. 4920--4928.

\bibitem{chen2017using}
S.~Chen, C.~Zhang, M.~Dong, J.~Le, M.~Rao, Using ranking-cnn for age
  estimation, in: Proceedings of the IEEE Conference on Computer Vision and
  Pattern Recognition, 2017, pp. 5183--5192.

\bibitem{liu2018constrained}
Y.~Liu, A.~Wai Kin~Kong, C.~Keong~Goh, A constrained deep neural network for
  ordinal regression, in: Proceedings of the IEEE Conference on Computer Vision
  and Pattern Recognition, 2018, pp. 831--839.

\bibitem{fu2018deep}
H.~Fu, M.~Gong, C.~Wang, K.~Batmanghelich, D.~Tao, Deep ordinal regression
  network for monocular depth estimation, in: Proceedings of the IEEE
  Conference on Computer Vision and Pattern Recognition, 2018, pp. 2002--2011.

\bibitem{herbrich1999support}
R.~Herbrich, T.~Graepel, K.~Obermayer, Support vector learning for ordinal
  regression (1999).

\bibitem{crammer2002pranking}
K.~Crammer, Y.~Singer, Pranking with ranking, in: Advances in neural
  information processing systems, 2002, pp. 641--647.

\bibitem{cao2019rank}
W.~Cao, V.~Mirjalili, S.~Raschka, Rank-consistent ordinal regression for neural
  networks, arXiv preprint arXiv:1901.07884 (2019).

\bibitem{fuchs2017spoken}
T.~Fuchs, J.~Keshet, Spoken term detection automatically adjusted for a given
  threshold, IEEE Journal of Selected Topics in Signal Processing 11~(8) (2017)
  1310--1317.

\bibitem{chu2007support}
W.~Chu, S.~S. Keerthi, Support vector ordinal regression, Neural computation
  19~(3) (2007) 792--815.

\bibitem{chu2005gaussian}
W.~Chu, Z.~Ghahramani, Gaussian processes for ordinal regression, Journal of
  machine learning research 6~(Jul) (2005) 1019--1041.

\bibitem{librosa}
B.~McFee, M.~McVicar, S.~Balke, V.~Lostanlen, C.~Thomé, C.~Raffel, D.~Lee,
  K.~Lee, O.~Nieto, F.~Zalkow, et~al.,
  \href{https://zenodo.org/record/2564164}{librosa/librosa: 0.6.3} (Feb 2019).
\newblock \href {https://doi.org/10.5281/zenodo.2564164}
  {\path{doi:10.5281/zenodo.2564164}}.
\newline\urlprefix\url{https://zenodo.org/record/2564164}

\bibitem{simonyan2014very}
K.~Simonyan, A.~Zisserman, Very deep convolutional networks for large-scale
  image recognition, arXiv preprint arXiv:1409.1556 (2014).

\bibitem{palermo2012dating}
F.~Palermo, J.~Hays, A.~A. Efros, Dating historical color images, in: European
  Conference on Computer Vision, Springer, 2012, pp. 499--512.

\bibitem{graham2015kaggle}
B.~Graham, Kaggle diabetic retinopathy detection competition report, University
  of Warwick (2015).

\bibitem{avaref}
N.~Murray, L.~Marchesotti, F.~Perronnin, Ava: A large-scale database for
  aesthetic visual analysis.

\bibitem{zhu2020deep}
H.~Zhu, Y.~Zhang, H.~Shan, L.~Che, X.~Xu, J.~Zhang, J.~Shi, F.-Y. Wang, Deep
  ordinal regression forests, arXiv preprint arXiv:2008.03077 (2020).

\bibitem{zhifei2017cvpr}
Z.~Zhang, Y.~Song, H.~Qi, Age progression/regression by conditional adversarial
  autoencoder, in: IEEE Conference on Computer Vision and Pattern Recognition
  (CVPR), IEEE, 2017.

\bibitem{leonard1993tidigits}
R.~G. Leonard, G.~Doddington, Tidigits speech corpus, Texas Instruments, Inc
  (1993).

\bibitem{Rothe-IJCV-2018}
R.~Rothe, R.~Timofte, L.~V. Gool, Deep expectation of real and apparent age
  from a single image without facial landmarks, International Journal of
  Computer Vision 126~(2-4) (2018) 144--157.

\bibitem{Rothe-ICCVW-2015}
R.~Rothe, R.~Timofte, L.~V. Gool, Dex: Deep expectation of apparent age from a
  single image, in: IEEE International Conference on Computer Vision Workshops
  (ICCVW), 2015.

\end{thebibliography}
